\documentclass[runningheads]{llncs}
\usepackage[T1]{fontenc}
\usepackage{graphicx}
\usepackage{amsmath}
\usepackage{amssymb}
\usepackage{siunitx}
\usepackage{booktabs}
\usepackage[table,xcdraw]{xcolor}
\begin{document}
\title{qMRI Diffuser: Quantitative T1 Mapping of the Brain using a Denoising Diffusion Probabilistic Model}
\titlerunning{qMRI Diffuser: Quantitative T1 Mapping using DDPM}
%
%
%
\author{Shishuai Wang \inst{1} \and 
Hua Ma \inst{1} \and 
Juan A. Hernandez-Tamames \inst{1,2} \and 
Stefan Klein \inst{1} \and 
Dirk H.J. Poot \inst{1}}
\authorrunning{S. Wang et al.}
%
%
\institute{Department of Radiology and Nuclear Medicine, Erasmus MC, Rotterdam, The Netherlands \and
Department of Imaging Physics, TU Delft, Delft, The Netherlands \\
\email{s.wang@erasmusmc.nl}}
\maketitle              

\begin{abstract}
Quantitative MRI (qMRI) offers significant advantages over weighted images by providing objective parameters related to tissue properties. 
Deep learning-based methods have demonstrated effectiveness in estimating quantitative maps from series of weighted images. 
In this study, we present qMRI Diffuser, a novel approach to qMRI utilising deep generative models.
Specifically, we implemented denoising diffusion probabilistic models (DDPM) for T1 quantification in the brain, framing the estimation of quantitative maps as a conditional generation task.
The proposed method is compared with the residual neural network (ResNet) and the recurrent inference machine (RIM) on both phantom and \textit{in vivo} data.
The results indicate that our method achieves improved accuracy and precision in parameter estimation, along with superior visual performance.
Moreover, our method inherently incorporates stochasticity, enabling straightforward quantification of uncertainty.
Hence, the proposed method holds significant promise for quantitative MR mapping.
\end{abstract}
%
%
%

\section{Introduction}
MRI is predominantly utilised for acquiring qualitative weighted images, which provide information regarding the contrast between various tissue types and are employed for visual assessment. 
However, compared with weighted images, quantitative MRI (qMRI) offers distinct advantages by measuring tissue properties, such as relaxation times and proton density.
As the tissue properties are objective and intrinsic, they exhibit the promise as biomarkers for various diseases \cite{cashmore2021clinical}.
Furthermore, qMRI enables the direct comparisons of data acquired under diverse conditions, e.g., different patients, time periods, scanning protocols, and hardware configurations.

The generation of quantitative maps typically involves the acquisition of a series of weighted images with distinct settings (e.g. different inversion times), from which the quantitative parameters can be estimated \cite{riwaj_review,qmri_mapping}.
Traditionally, the maximum likelihood estimator (MLE) \cite{og_mle} is used to estimate quantitative maps by iteratively fitting a physical signal model to the weighted images. 
However, MLE can suffer from high variability in scans with low signal-to-noise ratio (SNR) and requires substantial expertise for designing its regularisation term.
Alternatively, convolutional neural networks (CNN) have been trained to directly translate weighted images into quantitative maps \cite{cai2018single,shao2020fast,jeelani2020myocardial} and have proven effective.
However, they suffer from low interpretability and may fail without notification.
To leverage the advantages of both model-based and data-driven approaches, the recurrent inference machine (RIM) was proposed to iteratively reconstruct quantitative maps by training a recurrent neural network to predict the incremental update of each iteration step, achieving the state-of-the-art performance on T1 and T2 mapping tasks \cite{rim}.
However, this approach becomes computationally expensive for tasks with complicated (e.g. Bloch simulation-based) physical signal models, as it requires evaluation of the derivative of the likelihood with respect to the tissue parameters.

As a class of deep generative models, diffusion models have gained popularity by its impressive image generation capability.
There is also a burgeoning interest in applying diffusion models within the realm of medical image analysis, encompassing various tasks such as image generation, reconstruction, registration, classification, segmentation, denoising and detection \cite{diff_review}.
Inspired by the success of diffusion models, in this work, we present qMRI Diffuser and approach quantitative MR mapping as a generation task performed by a denoising diffusion probabilistic model (DDPM) \cite{ddpm} to take the advantage of diffusion models' image generation quality.
To the best of our knowledge, this is the first use of diffusion model for qMRI.
In the proposed approach, the generation of quantitative maps starts with a Gaussian noise, and the results are obtained by the gradual denoising using a deep learning model.
To estimate the quantitative parameters (e.g. T1) corresponding to a given series of weighted images, we set the weighted images as the condition that guides the generation process.
In addition, given the inherent stochastic nature of the proposed method, uncertainty can be readily quantified by repeating the inference process \cite{sbdmmri}.

To investigate the advantages of our proposed method in qMRI, we compare it with two established deep learning-based methods: 1) ResNet as a baseline, and 2) the RIM-based method \cite{rim}, in a T1 mapping task of the brain.
To address the lack of training data, we generated a synthetic dataset for the training of the three methods, while the evaluation was conducted using real scan data acquired from a hardware phantom and a healthy volunteer.
Compared with ResNet and RIM, we found that the proposed method exhibits better accuracy and precision as well as superior visual performance.
Moreover, our proposed method provides meaningful uncertainty information without specific treatments, which is particularly useful.
To summarise, the following contributions are made in this paper:
\begin{itemize}
    \item To the best of our knowledge, this is the first diffusion model-based method developed for qMRI applications
    \item Improved visual performance and parameter estimation quality compared to other deep learning-based methods, as showcased by phantom and \textit{in vivo} testing data
    \item Improved trustworthiness and interpretability of estimation results by the straightforward uncertainty quantification ability of the proposed method
\end{itemize}

\section{Method}
\begin{figure}[h]
    \centering
    \includegraphics[width=0.75\textwidth]{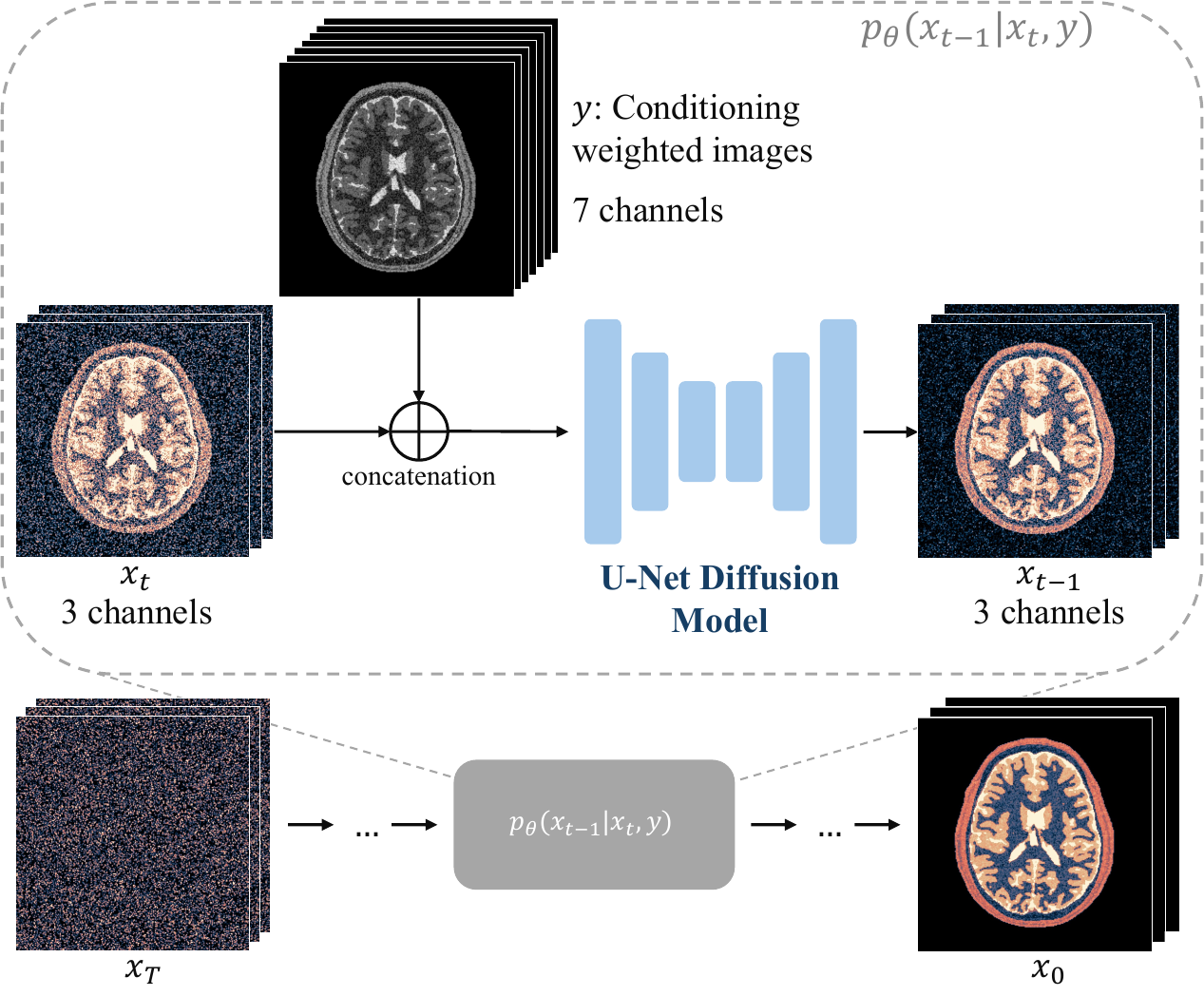}
    \caption{Illustration of qMRI Diffuser's workflow. A U-Net model is trained to predict the noise added at a given time step. Weighted images acquired at different contrast states are set as the condition for the generation of quantitative maps. The number of channels in $y$ and $x_t$ should correspond to the number of weighted images used and quantitative parameters of interest respectively, and specifically here we set them as 7 and 3.}
    \label{qmridiff}
\end{figure}

DDPM defines a forward diffusion process and the reverse process.
Assuming $x_0$ represents the clean original data sample, the forward process gradually introduces Gaussian noise $\epsilon$ to $x_0$ through a scheduler over a series of $T$ steps, thereby reshaping the original distribution of $x_0$ into a Gaussian distribution of $x_T$.
The reverse process starts from $x_T$, which is Gaussian noise, and uses a model parametrised by $\theta$ to reverse the noise adding steps as \cite{ddpm,diff_review,conditional}
\begin{equation}
    p_\theta\left(x_{0: T} \mid y\right)=p\left(x_T\right) \prod_{t=1}^T p_\theta\left(x_{t-1} \mid x_t, y\right)
    \label{ddpm_rev}
\end{equation}
where $y$ is the condition to guide the reverse (i.e. generation) process. 
Specifically, in our task, we set $x_0$ as the clean quantitative maps, where the number of channels corresponds to the number of quantitative parameters to estimate, and we use weighted images as the condition.
The quantification task is then executed through a series of the reverse processes $p_\theta\left(x_{t-1} \mid x_t,y\right)$, where $t$ starts from $T$ and decreases to 1, ultimately producing quantitative maps corresponding to the respective conditioning weighted images.
The training of the model is conducted by minimising the following objective function \cite{ddpm}
\begin{equation}
    L=\mathbb{E}_{t, x_0, \epsilon}\left[\left\|\epsilon-\epsilon_\theta\left(x_t, y, t\right)\right\|^2\right]
    \label{ddpm_loss}
\end{equation}
where $\epsilon_\theta\left(x_t, y, t\right)$ is trained to approximate $\epsilon$ given the time step $t$, the noisy sample $x_t$, and the condition $y$ as inputs.
Here, we provide $y$ to the model by concatenating it with $x_t$ in the channel dimension.
The workflow of parameter mapping using the trained model is illustrated in Fig. \ref{qmridiff}.

\section{Experiments}
\subsection{Scanning Protocol}
\label{Scanning Protocol}
As a proof of concept for the proposed method, we deployed a inversion recovery fast spin echo (IR-FSE) protocol on a GE M750 scanner with a 16-channel head and neck coil to acquire real scan data.
For each scan, 30 axial slices were acquired with a voxel size of $1\times1\times2 mm^3$.
We used TR = 10 seconds, TE = 7.14 ms, ETL = 64, FOV $= [250, 250, 60]$ mm, and acquisition matrix $= [240, 240]$.
The inversion times (TI) were set as $[0.05, 0.10, 0.25, 0.50, 0.85, 1.50, 2.50]$ seconds to get seven weighted images at different contrast states, which were used for quantitative parameter mapping.

With the selected protocol, there are three quantitative parameters to estimate: T1, proton density and inversion efficiency, with T1 being of the most interest.
Although inversion efficiency is mostly related to radio-frequency (RF) and does not directly reflect tissue properties, it could confound the T1 estimation.
Therefore, we are interested in estimating it as well.

\subsection{Datasets}
The training dataset was synthesised following \cite{rim} and consists of paired quantitative maps and weighted images.
The anatomy structure of quantitative maps were provided by 3D virtual discrete brain models from BrainWeb \cite{cocosco1997brainweb}, and the parameter values were drawn from the realistic distributions of a given tissue type \cite{whatisT1T2}.
Subsequently, a series of weighted images were generated based on the synthetic quantitative maps and a forward signal model of the IR-FSE acquisition detailed in Section \ref{Scanning Protocol}:
\begin{equation}
    S_n=\left|PD\left(1-B e^{-\frac{{TI}_n}{T_1}}\right)\right|
    \label{forward_model}
\end{equation}
where $n$ is the index of weighed images, $S_n \in \left \{ S_{1},...,S_{7}  \right \}$ represents the weighted image at inversion time ${TI}_n$. 
$PD$ and $B$ denotes proton density and inversion efficiency, respectively.
In total 994 geometrically different axial slices were drawn from 14 brain models.
For each slice, 4 realisations of quantitative parameter combination were generated, forming a training dataset with 3976 slices.

The testing dataset were acquired from: 1) a quantitative NIST/ISMRM phantom (CaliberMRI, Boulder, CO, USA), and 2) a healthy volunteer, with approval from the our institutional review board and informed consent from the volunteer obtained.

\subsection{Network training}
The architecture and hyperparameters of ResNet and RIM were set as the same as in \cite{rim}, except that we increased RIM's number of inference steps from 6 to 9 to boost its performance. 
The qMRI Diffuser was deployed based on the open-source platform MONAI Generative Models \cite{monai}.
Within the qMRI Diffuser, a U-Net with time step embedding \cite{diff_unet1,diff_unet2} was configured, comprising three levels with the number of channels set to 128, 256, and 256, respectively.
The original DDPM noise scheduler was used \cite{ddpm}, which means the expected range of quantitative maps is $[-1.0, 1.0]$.
Hence, we scaled the training quantitative maps during pre-processing stage using $f(\boldsymbol{x})=2\times tanh(\boldsymbol{x})-1$.
After training, the predicted quantitative maps were scaled back using the inverse of this function.
All three methods were trained by using the same synthetic dataset with batch size of 8 and 100 epochs.

\subsection{Evaluation}
The phantom data and \textit{in vivo} data were used to evaluate and compare ResNet, RIM and the proposed qMRI Diffuser.
For qMRI Diffuser, we repeated the inference 10 times with different randomly initialised $x_{T}$ and took the mean of the outputs as the final result.
Meanwhile, the voxel-wise standard deviation of the 10 outputs can act as a measurement of uncertainty.
The evaluation focused on the T1 and proton density (PD) channels, as inversion efficiency can not act as a biomarker.
Visual evaluation was conducted on both phantom data and \textit{in vivo} data, and T1 mapping results were also quantitatively evaluated using the phantom.

\section{Results}
\begin{figure}[!htb]
    \centering
    \includegraphics[width=\textwidth]{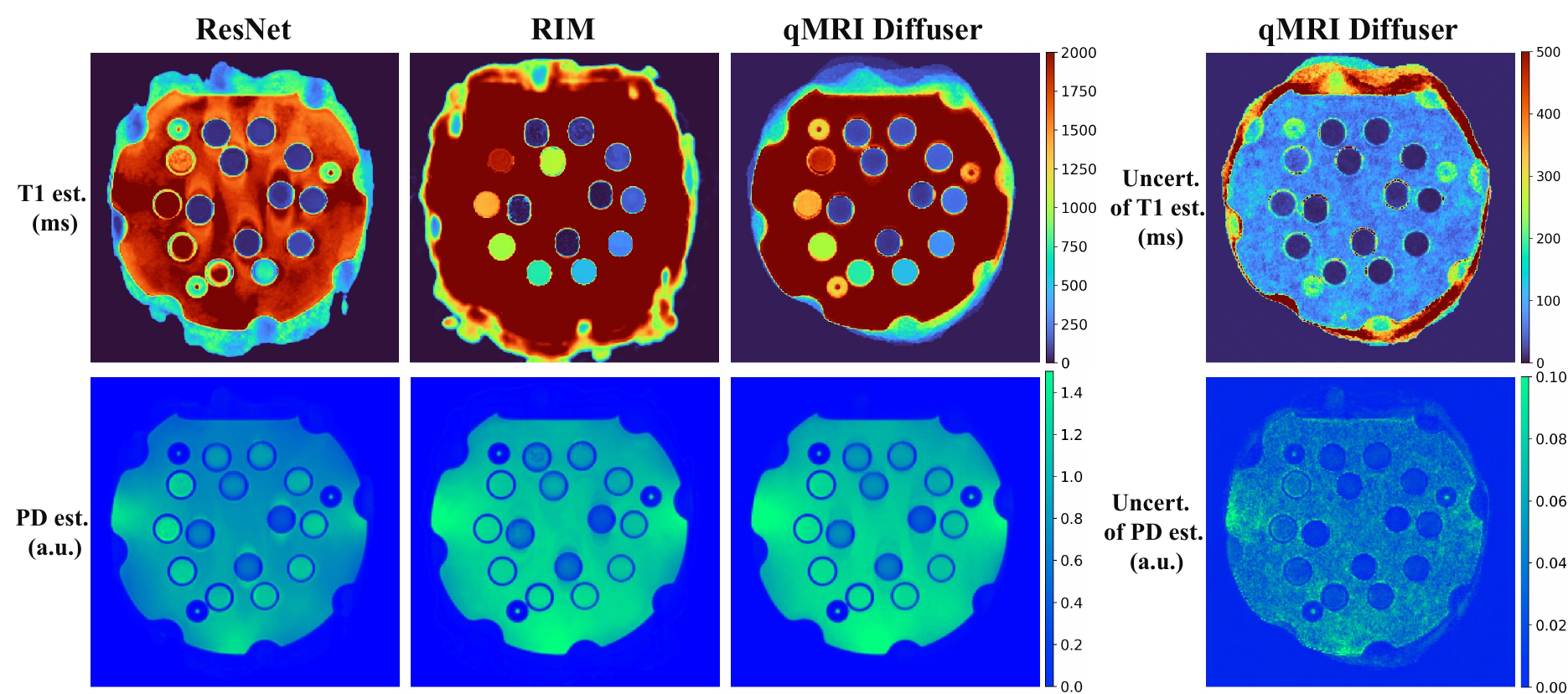}
    \caption{Example of T1 and PD estimation results in the phantom. A slice containing the $NiCl_2$ T1 spheres is presented.}
    \label{phantom_eva}
\end{figure}

Fig. \ref{phantom_eva} shows a representative example of the T1 and PD estimation results for the phantom.
With the used IR-FSE protocol, a typical artifact in weighted images is the blurring around the spheres. 
Similar visual effects are also observed in the T1 estimation from ResNet and PD estimations from all methods. 
For the T1 estimation from RIM, such artifacts did not directly cause blurring of the spheres but resulted in their distortion.
All three methods exhibit blurring around the phantom due to signal interference from the background, but the proposed method gives the clearest edge of the phantom.
The PD estimations are visually comparable among all three methods, except that the estimation from ResNet is relatively lower in the water bolus.

The mean and standard deviation of T1 estimations in each T1-sphere are summarised in Table \ref{T1quantitative}.
Within the relevant T1 range typically found \textit{in vivo} (i.e., from 400 to 2000 ms), the mean value of T1 estimations using qMRI Diffuser has the closest match to the ground truth values.
Additionally, these estimations exhibit the smallest standard deviation in the ROIs.
For T1 lower than 500 ms, qMRI Diffuser exhibits a bias towards values larger than the ground truth, with a standard deviation mostly smaller than that of RIM but larger than that of ResNet.

\begin{table}[!htb]
\centering
\caption{Quantitative evaluation of T1 mapping (ms) on the phantom at room temperature.
Ground truth (GT) values are from the manual (properties of T1 contrast spheres at 3T, \SI{20}{\degreeCelsius}). 
The results for spheres with T1 values within the typical \textit{in vivo} range are highlighted in bold.}
\label{T1quantitative}
\begin{tabular}{@{}cccccccc@{}}
\toprule
\textbf{GT} &
  \textbf{ResNet} &
  \textbf{RIM} &
  \textbf{Diffuser} &
  \textbf{GT} &
  \textbf{ResNet} &
  \textbf{RIM} &
  \textbf{Diffuser} \\ \midrule
\textbf{\begin{tabular}[c]{@{}c@{}}1884\\ $\pm$30\end{tabular}} &
  \textbf{\begin{tabular}[c]{@{}c@{}}1534 \\ $\pm$72\end{tabular}} &
  \textbf{\begin{tabular}[c]{@{}c@{}}1835 \\ $\pm$39\end{tabular}} &
  \textbf{\begin{tabular}[c]{@{}c@{}}1772 \\ $\pm$36\end{tabular}} &
  \begin{tabular}[c]{@{}c@{}}175\\ $\pm$3\end{tabular} &
  \begin{tabular}[c]{@{}c@{}}121\\ $\pm$13\end{tabular} &
  \begin{tabular}[c]{@{}c@{}}186 \\ $\pm$23\end{tabular} &
  \begin{tabular}[c]{@{}c@{}}224\\ $\pm$14\end{tabular} \\
\textbf{\begin{tabular}[c]{@{}c@{}}1330\\ $\pm$20\end{tabular}} &
  \textbf{\begin{tabular}[c]{@{}c@{}}4575\\ $\pm$1203\end{tabular}} &
  \textbf{\begin{tabular}[c]{@{}c@{}}1365 \\ $\pm$22\end{tabular}} &
  \textbf{\begin{tabular}[c]{@{}c@{}}1353 \\ $\pm$20\end{tabular}} &
  \begin{tabular}[c]{@{}c@{}}121\\ $\pm$2\end{tabular} &
  \begin{tabular}[c]{@{}c@{}}124\\ $\pm$12\end{tabular} &
  \begin{tabular}[c]{@{}c@{}}101\\ $\pm$31\end{tabular} &
  \begin{tabular}[c]{@{}c@{}}176\\ $\pm$15\end{tabular} \\
\textbf{\begin{tabular}[c]{@{}c@{}}987\\ $\pm$14\end{tabular}} &
  \textbf{\begin{tabular}[c]{@{}c@{}}2452\\ $\pm$418\end{tabular}} &
  \textbf{\begin{tabular}[c]{@{}c@{}}966\\ $\pm$17\end{tabular}} &
  \textbf{\begin{tabular}[c]{@{}c@{}}1003\\ $\pm$15\end{tabular}} &
  \begin{tabular}[c]{@{}c@{}}85\\ $\pm$1\end{tabular} &
  \begin{tabular}[c]{@{}c@{}}89\\ $\pm$10\end{tabular} &
  \begin{tabular}[c]{@{}c@{}}90\\ $\pm$140\end{tabular} &
  \begin{tabular}[c]{@{}c@{}}161\\ $\pm$25\end{tabular} \\
\textbf{\begin{tabular}[c]{@{}c@{}}690\\ $\pm$10\end{tabular}} &
  \textbf{\begin{tabular}[c]{@{}c@{}}1654\\ $\pm$415\end{tabular}} &
  \textbf{\begin{tabular}[c]{@{}c@{}}683\\ $\pm$18\end{tabular}} &
  \textbf{\begin{tabular}[c]{@{}c@{}}695\\ $\pm$16\end{tabular}} &
  \begin{tabular}[c]{@{}c@{}}60\\ $\pm$1\end{tabular} &
  \begin{tabular}[c]{@{}c@{}}78\\ $\pm$11\end{tabular} &
  \begin{tabular}[c]{@{}c@{}}1053\\ $\pm$62\end{tabular} &
  \begin{tabular}[c]{@{}c@{}}139\\ $\pm$21\end{tabular} \\
\textbf{\begin{tabular}[c]{@{}c@{}}485\\ $\pm$7\end{tabular}} &
  \textbf{\begin{tabular}[c]{@{}c@{}}617\\ $\pm$139\end{tabular}} &
  \textbf{\begin{tabular}[c]{@{}c@{}}504\\ $\pm$15\end{tabular}} &
  \textbf{\begin{tabular}[c]{@{}c@{}}503\\ $\pm$13\end{tabular}} &
  \begin{tabular}[c]{@{}c@{}}43\\ $\pm$1\end{tabular} &
  \begin{tabular}[c]{@{}c@{}}60\\ $\pm$9\end{tabular} &
  \begin{tabular}[c]{@{}c@{}}68\\ $\pm$98\end{tabular} &
  \begin{tabular}[c]{@{}c@{}}116\\ $\pm$19\end{tabular} \\
\begin{tabular}[c]{@{}c@{}}342\\ $\pm$5\end{tabular} &
  \begin{tabular}[c]{@{}c@{}}175\\ $\pm$62\end{tabular} &
  \begin{tabular}[c]{@{}c@{}}354 \\ $\pm$16\end{tabular} &
  \begin{tabular}[c]{@{}c@{}}370\\ $\pm$14\end{tabular} &
  \begin{tabular}[c]{@{}c@{}}30\\ $\pm$1\end{tabular} &
  \begin{tabular}[c]{@{}c@{}}66\\ $\pm$9\end{tabular} &
  \begin{tabular}[c]{@{}c@{}}26\\ $\pm$18\end{tabular} &
  \begin{tabular}[c]{@{}c@{}}100\\ $\pm$19\end{tabular} \\
\begin{tabular}[c]{@{}c@{}}241\\ $\pm$3\end{tabular} &
  \begin{tabular}[c]{@{}c@{}}139\\ $\pm$26\end{tabular} &
  \begin{tabular}[c]{@{}c@{}}211\\ $\pm$18\end{tabular} &
  \begin{tabular}[c]{@{}c@{}}263\\ $\pm$12\end{tabular} &
  \begin{tabular}[c]{@{}c@{}}21\\ $\pm$1\end{tabular} &
  \begin{tabular}[c]{@{}c@{}}117\\ $\pm$31\end{tabular} &
  \begin{tabular}[c]{@{}c@{}}34\\ $\pm$41\end{tabular} &
  \begin{tabular}[c]{@{}c@{}}118\\ $\pm$31\end{tabular} \\ \bottomrule
\end{tabular}
\end{table}

\begin{figure}[!htb]
    \centering
    \includegraphics[width=\textwidth]{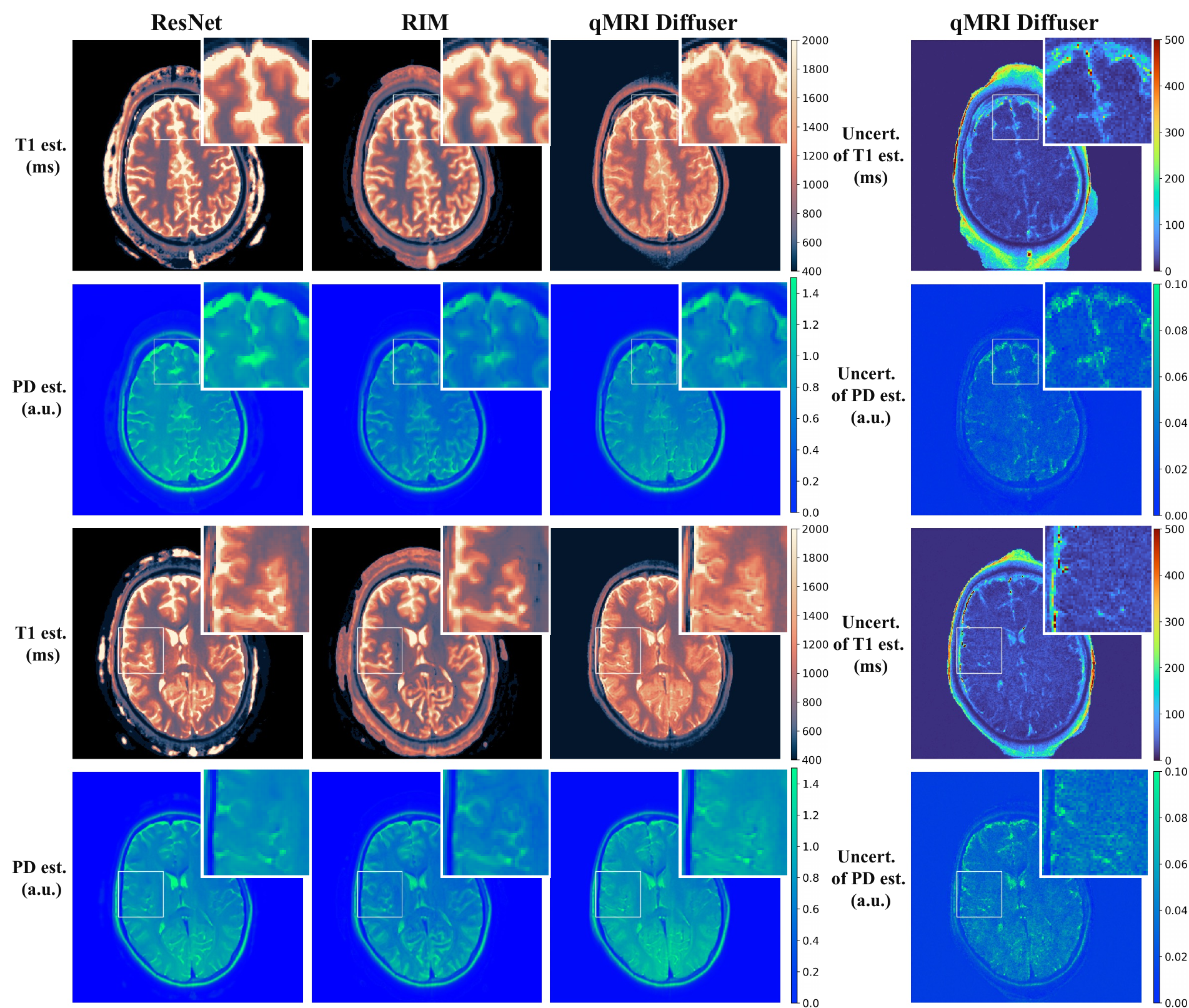}
    \caption{Example of T1 and PD estimation results on the \textit{in vivo} data. The \textit{in vivo} T1 maps are plotted following the colormap recommendations \cite{cmap}.}
    \label{invivo_eva}
\end{figure}

Fig. \ref{invivo_eva} shows examples of T1 and PD estimations for \textit{in vivo} data.
In the T1 estimation, qMRI Diffuser shows the ability to preserve small anatomical structures, while blurriness around small structures are observed from both ResNet's and RIM's results.
Within the brain mask, high uncertainty values given by qMRI Diffuser are observed in the high T1 region (above 2000 ms), while the uncertainty values in gray matter/white matter region is approximately homogeneous.
The PD estimations from all methods are visually similar without outliers or noticeable artifacts, while RIM's estimation is lower in the white matter and qMRI Diffuser's estimation is slightly higher overall.

\section{Discussion}
According to the evaluation results, the proposed method trained purely on synthetic data generalises well to data from real scans, and shows resilience to artifacts in the input weighted images.
The test with phantom shows that qMRI Diffuser achieves the best visual results for T1 mapping by providing clean edges of spheres and uniform T1 values within the spheres, as well as higher accuracy and precision for T1 values in the range typically found \textit{in vivo}.
Though qMRI Diffuser's performance for T1 values less than 500 ms is admittedly suboptimal, this could be attributed to the fact that such T1 values are rare in the synthetic human brain training dataset, and the fact that qMRI Diffuser is not physics model-based, which may limit the ability to extrapolate outside of the range of values seen during training.
However, it could be improved by using a training dataset with a larger T1 range.
The proposed method also provides superior visual performance on the \textit{in vivo} data compared to ResNet and RIM, and the estimated T1 maps using qMRI Diffuser accurately depicted anatomical structures including the fine details.

One significant advantage of the proposed method is that it readily allows uncertainty quantification, which adds trustworthiness and interpretability to this method.
For other methods, uncertainty quantification is either infeasible or dependent on separate operations (e.g. by adding dropout layers in the neural network \cite{gal2016dropout}).
The uncertainty quantification by using voxel-wise standard deviation of repeated inferences is meaningful, as the result reconstructed from a randomly initialised $x_{t}$ can be viewed as a sample from the posterior distribution, given the training data the model has seen.
It is noticeable that high-uncertainty region given by qMRI Diffuser overlaps with its error-prone areas (e.g. edge of spheres, high T1 regions), indicating the meaningfulness and usefulness of the uncertainty information provided by our method.

The proposed method effectively addresses the challenge of limited weighted images as demonstrated in the experiments, aligning with the objective of reducing the scanning time to acquire quantitative maps.
Compared to RIM which requires derivative computations through the signal model, the proposed method can be applied to other scanning protocols with more complicated signal models without introducing additional challenges or computational expenses as it does not require such computations, enhancing its usability across a wide range of scanning protocols.

\section{Conclusion}
We propose a novel quantitative MR mapping method, qMRI Diffuser, which frames the mapping process as a conditional generation task using a denoising diffusion probabilistic model (DDPM).
The visual and quantitative evaluation on real scans demonstrates the superior performance of the proposed method, highlighting its potential in clinical use.
Particularly, the proposed method allows readily uncertainty quantification.
In conclusion, the proposed method could be a promising tool for qMRI.

\begin{credits}
\subsubsection{\ackname} This work was conducted within the “Trustworthy AI for MRI” ICAI lab within the project ROBUST, funded by the Dutch Research Council (NWO), GE Healthcare, and the Dutch Ministry of Economic Affairs and Climate Policy (EZK).

\subsubsection{\discintname}
Shishuai Wang, Stefan Klein, Juan-Antonio Hernandez-Tamames and Dirk Poot received research grants from GE Healthcare. Hua Ma has no competing interests related to the scope of this article.
\end{credits}

%
%
%
\bibliographystyle{splncs04}
\bibliography{Paper-0018}

\end{document}